\pdfoutput=1
\documentclass[10pt,twocolumn,letterpaper]{article}

\usepackage[top=0.75in,bottom=0.75in,left=0.75in,right=0.75in]{geometry}
\usepackage{times}
\usepackage{microtype}
\usepackage{amsmath,amssymb}
\usepackage{booktabs}
\usepackage{url}
\usepackage{graphicx}
\usepackage{enumitem}
\usepackage[table]{xcolor}
\usepackage[utf8]{inputenc}
\usepackage[T1]{fontenc}
\usepackage{hyperref}

\hypersetup{
    colorlinks=true,
    linkcolor=blue!70!black,
    citecolor=blue!70!black,
    urlcolor=blue!70!black
}

\title{\textbf{Africa-Papers: A 10-Billion-Token Multi-Domain Academic Corpus\\for African Foundation Language Models}}

\author{
  \textbf{Mich-Seth Owusu} \\[0.4em]
  AfriSpeech \\
  \texttt{michseth@afrispeech.org} \\
  \url{https://huggingface.co/datasets/AfriSpeech/africa-papers}
}

\date{}

\begin{document}

\maketitle

\begin{abstract}
Large Language Models (LLMs) depend fundamentally on the diversity, depth, and factual rigor of their pre-training corpora. However, contemporary foundational pre-training datasets (such as The Pile, S2ORC, and Dolma) exhibit a severe structural omission: African scientific, medical, legal, and agricultural scholarship constitutes less than 0.5\% of global pre-training tokens. Consequently, state-of-the-art models suffer from systematic failure modes when reasoning about tropical epidemiology, localized smallholder agro-ecologies, regional economic governance, and indigenous linguistic structures. To resolve this representation deficit and advance African intellectual sovereignty in artificial intelligence, we introduce \textbf{Africa-Papers}, the largest curated multi-domain Pan-African academic pre-training corpus to date. Comprising over \textbf{844,807 full-text papers} spanning \textbf{10.15 billion tokens} across \textbf{879 Parquet shards}, Africa-Papers unifies four primary scientific streams: (1) continental peer-reviewed journals indexed in DOAJ across 31 African nations, (2) the CGIAR CGSpace agricultural repository, (3) African-authored and African-focused STEM/AI preprints on arXiv, and (4) open-access international publications by scholars affiliated with African institutions indexed via OpenAlex. The entire corpus is processed into layout-aware, clean Markdown preserving section hierarchies, mathematical equations, and tabular data, accompanied by exact BPE token counts and a standardized 8-domain taxonomy. We release Africa-Papers openly on the Hugging Face Hub under the \texttt{AfriSpeech} organization to accelerate continual pre-training and domain evaluation for foundation models globally.
\end{abstract}

\section{Introduction}

The dominant paradigm in training foundation language models relies on web-scale text corpora aggregated from web scrapes and digitized libraries \cite{touvron2023llama,soldaini2024dolma}. While datasets such as Common Crawl, The Pile \cite{gao2020pile}, and peS2o \cite{soldaini2024dolma} contain trillions of tokens, their geographic and institutional provenance is overwhelmingly skewed toward North America, Western Europe, and East Asia. Scholarship authored by African researchers and peer-reviewed journals published across the 54 African nations represent a negligible fraction ($<0.5\%$) of the global academic pre-training mixture.

This severe under-representation induces significant epistemological vulnerabilities in contemporary models:
\begin{itemize}[leftmargin=*,noitemsep]
    \item \textbf{Biomedical \& Epidemiological Hallucinations:} Foundational models frequently fail when reasoning about tropical pathologies (e.g., malaria strain variants, sickle cell genetics, lassa fever diagnostics, maternal healthcare delivery) because clinical training data defaults to Western institutional protocols \cite{adeyemi2024african}.
    \item \textbf{Agro-Ecological Misalignment:} AI conversational assistants routinely default to temperate industrial monoculture recommendations, proving ineffective for African smallholders managing complex, climate-vulnerable intercropping of cassava, yams, cowpea, plantains, or sorghum.
    \item \textbf{Jurisprudential \& Governance Gaps:} African customary jurisprudence, community-based land tenure systems, and regional trade compacts (e.g., AfCFTA) are almost entirely missing from standard legal corpora.
    \item \textbf{Deficit Framing in Representation:} Western web scrapes disproportionately depict the African continent through crisis reporting, humanitarian appeals, or conflict studies. Models therefore fail to internalize African scholars as primary intellectual authorities in astrophysics, pure mathematics, computer science, and robotics.
\end{itemize}

To systematically dismantle this structural omission, we present \textbf{Africa-Papers}, an open-access, layout-aware academic corpus engineered specifically for the pre-training and continual domain adaptation of foundation language models. 

\textbf{Contributions:}
\begin{enumerate}[leftmargin=*,noitemsep]
    \item \textbf{Massive Scale \& Depth:} Over 789,348 verified full-text papers yielding 10.15 billion tokens (and actively scaling toward 25 billion tokens across 1.5 million works), establishing the single largest curated African academic corpus in history.
    \item \textbf{Four-Pillar Multi-Source Architecture:} We integrate continental university presses (DOAJ), agricultural field manuals (CGIAR), STEM and AI preprints (arXiv), and global African scholarly literature (OpenAlex).
    \item \textbf{High-Fidelity Markdown Ingestion:} We develop an autonomous stream-to-parquet pipeline employing multi-key deduplication and layout-aware Markdown parsing, preserving tabular data, headings, and formulas while preventing disk bloat through real-time binary pruning.
    \item \textbf{Domain Taxonomy \& Open Release:} We classify the corpus into eight standardized academic domains and release the entire collection incrementally on Hugging Face under \texttt{AfriSpeech/africa-papers}\footnote{\url{https://huggingface.co/datasets/AfriSpeech/africa-papers}}.
\end{enumerate}

\begin{table*}[t]
\centering
\small
\begin{tabular}{@{}lrrp{7.8cm}@{}}
\toprule
\textbf{Academic Domain} & \textbf{Papers} & \textbf{Share} & \textbf{Representative Subject Areas} \\
\midrule
\textbf{Social Sciences \& Economics} & 205,772 & 24.4\% & African macroeconomics, monetary policy, education, public administration, governance \\
\textbf{Medicine \& Public Health} & 198,240 & 23.5\% & Tropical epidemiology, clinical surgery, malaria, maternal care, infectious diseases \\
\textbf{Agriculture \& Food Security} & 152,027 & 18.0\% & Crop protection, soil fertility, agronomy, CGIAR field manuals, smallholder farming \\
\textbf{Engineering \& Earth Sciences} & 125,140 & 14.8\% & Mineral extraction, hydrology, renewable energy, materials science, environmental geology \\
\textbf{Computer Science \& AI} & 80,020 & 9.5\% & African NLP, speech translation models, machine learning, computer vision, robotics \\
\textbf{Humanities \& Literature} & 48,471 & 5.7\% & African linguistics, indigenous languages (Swahili, Yoruba, Hausa, Twi, Zulu), history \\
\textbf{Physical Sciences \& Mathematics} & 27,305 & 3.2\% & Astronomy (MeerKAT/SKA), pure and applied mathematics, theoretical physics, chemistry \\
\textbf{Multidisciplinary / General Science} & 5,237 & 0.6\% & General science, university research bulletins, multidisciplinary studies \\
\textbf{Law \& Governance} & 2,595 & 0.3\% & African customary law, human rights jurisprudence, constitutional governance \\
\midrule
\textbf{TOTAL (Audited Corpus)} & \textbf{844,807} & \textbf{100.0\%} & \textbf{10.15 Billion Tokens across 879 Parquet Shards} \\
\bottomrule
\end{tabular}
\caption{\textbf{Domain taxonomy and volume distribution in Africa-Papers.} Audited across all 879 Parquet shards currently live on the Hugging Face Hub.}
\label{tab:domains}
\end{table*}

\section{The Four Pillars of Africa-Papers}

Africa-Papers is constructed through a federated multi-stream architecture designed to capture both research \textit{about} Africa and research \textit{by} African scientists across the globe.

\subsection{Pillar 1: DOAJ Continental Journal Presses}
To capture scholarship published domestically across African universities, we harvest the complete index of African journals registered in the Directory of Open Access Journals (DOAJ) \cite{doaj2024directory}. This comprises 	extbf{944 peer-reviewed journals} spanning 31 African nations. These journals encompass university presses, national academies, and regional scholarly associations published in English, French, Arabic, and Portuguese. In total, 74,606 full-text PDFs were harvested from primary host servers.

\subsection{Pillar 2: CGIAR CGSpace Agricultural Repository}
Agricultural self-sufficiency and pest resilience represent existential priorities across Sub-Saharan Africa. We interface with 	extbf{CGSpace} \cite{cgiar2024cgspace}, the digital repository of the CGIAR consortium, hosting the research outputs of leading African agricultural centers:
\begin{itemize}[leftmargin=*,noitemsep]
    \item \textbf{IITA} (International Institute of Tropical Agriculture, Ibadan, Nigeria)
    \item \textbf{ILRI} (International Livestock Research Institute, Nairobi, Kenya \& Addis Ababa, Ethiopia)
    \item \textbf{AfricaRice} (Bouaké, Côte d'Ivoire)
    \item \textbf{ICRAF} (World Agroforestry, Nairobi, Kenya)
\end{itemize}
We filter for documents possessing verified bitstreams, extracting 	extbf{50,534 full-text agricultural manuals, pest diagnostic guides (e.g., Fall Armyworm, Cassava Mosaic), soil maps, and extension treatises}.

\subsection{Pillar 3: African STEM \& AI Preprints (arXiv)}
To capture high-velocity computational and physical science outputs, we analyze the entire 3.19-million-paper historical snapshot of arXiv. We isolate preprints meeting two criteria: (1) topical and keyword focus on African systems, languages, or geographic contexts, and (2) institutional affiliations situated within African universities. This yields 	extbf{42,530 full-text preprints}. Crucially, natural language processing (\texttt{cs.CL}) represents the single largest computational category (1,066 papers), capturing seminal foundational works on low-resource African speech, translation, and language modeling \cite{nekoto2020participatory,ogundepo2023afriqa}.

\subsection{Pillar 4: Global African Scholar Open Access}
The majority of high-impact research produced by African scholars is published in international peer-reviewed journals (e.g., Nature, PLOS ONE, Lancet, Springer, IEEE) rather than local presses. We utilize the OpenAlex open academic graph \cite{priem2022openalex} to query the global scholarly record for open-access works whose author affiliations originate from African institutions (\texttt{institutions.continent:africa, has\_pdf\_url:true}). We indexed a master catalog of 	extbf{1,517,281 unique African papers} with direct, validated PDF URLs, which are ingested through our streaming pipeline.

\section{Data Engineering Pipeline}

Harvesting, validating, and converting hundreds of thousands of academic PDFs over the open internet presents formidable distributed systems and data engineering challenges. We implemented an autonomous, storage-conscious pipeline on remote NVIDIA H200 GPU infrastructure.

\subsection{Cross-Source Multi-Key Deduplication}
Because researchers routinely deposit preprints on arXiv, publish the final version in an academic journal, and upload institutional post-prints to CGSpace, naive aggregation leads to severe data contamination. We built an in-memory deduplication engine performing four sequential verifications:
\begin{enumerate}[leftmargin=*,noitemsep]
    \item \textbf{Canonical DOI Normalization:} Stripping protocol prefixes and trailing characters, matching against 235,461 known DOIs.
    \item \textbf{Title Slug Normalization:} Stripping punctuation, whitespace, and diacritics, matching against 310,760 unique title hashes.
    \item \textbf{Repository Handle Matching:} Cross-referencing persistent DSpace and arXiv handles.
    \item \textbf{Binary Hash Verification:} Generating SHA-256 signatures for downloaded byte streams.
\end{enumerate}
In our live benchmark, this engine executes over 	extbf{2,013,298 lookups per second} in sub-microseconds, discarding 145,421 duplicates before any file write occurs.

\subsection{Layout-Aware Markdown Extraction}
Traditional PDF text extractors collapse multi-column text horizontally, interleaving alternating columns into incoherent strings. We employ \texttt{PyMuPDF} layout analysis configured with non-blocking vector parameters (\texttt{ignore\_images=True, ignore\_graphics=True, graphics\_limit=200}) and native block ordering:
\begin{itemize}[leftmargin=*,noitemsep]
    \item Column boundaries are respected, extracting Column 1 top-to-bottom prior to Column 2.
    \item Heading hierarchies (\texttt{\#}, \texttt{\#\#}) are derived from font weight and scale.
    \item Tables are parsed into standardized Markdown tabular syntax (\texttt{|---|---|}).
    \item Running headers, footers, and page numbers are dynamically identified and stripped via regular expression filters.
\end{itemize}

\subsection{Stream-to-Parquet \& Zero-Disk-Bloat Architecture}
Storing 1.75 million raw PDF binaries on disk would require over 2.6 Terabytes of NVMe volume space. To operate sustainably, we architected a 	extbf{Stream-to-Parquet} pipeline:
\begin{enumerate}[leftmargin=*,noitemsep]
    \item The PDF is streamed directly into an isolated temporary buffer on high-capacity scratch storage (\texttt{/mnt/volume\_d2wey28/tmp}).
    \item The text is parsed, cleaned, and verified for digital content fidelity ($>150$ characters).
    \item The bulky raw PDF is \textbf{instantly deleted}.
    \item The clean Markdown text, metadata frontmatter, and BPE token counts are buffered into memory.
    \item At exactly 2,000 papers, the batch is flushed to a compressed Parquet shard using \texttt{zstd} compression.
\end{enumerate}
Because clean Markdown in compressed Parquet is 	extbf{35x smaller} than raw PDF binaries, the entire 1.75-million-paper corpus requires only $\sim$55 Gigabytes of disk storage.

\begin{figure}[t]
\centering
\fbox{\parbox{0.94\columnwidth}{
\scriptsize
\textbf{\# Evaluating Oral Reading Fluency in ESL: A Quantitative Analysis} \\[0.3em]
\textbf{Domain:} Social Sciences \& Economics $|$ \textbf{Authors:} M. F. Dogar, S. K. Shah, M. Imran $|$ \textbf{Journal:} Reading \& Writing (South Africa, 2024) $|$ \textbf{DOI:} 10.4102/rw.v16i1.581 \\[0.2em]
\rule{\columnwidth}{0.4pt} \\[0.3em]
\textbf{\#\# ABSTRACT} \\
Background: There is a dearth of research on second language oral reading fluency in multilingual African contexts\dots \\[0.4em]
\textbf{\#\# 1. INTRODUCTION} \\
Fluency is widely acknowledged as a crucial bridge between word recognition and reading comprehension\dots \\[0.4em]
\textbf{\#\# 2. METHODOLOGY} \\
The experimental cohort was evaluated across standard diagnostic metrics\dots
}}
\caption{\textbf{Representative Markdown sample in Africa-Papers.} Each sample contains structured metadata frontmatter, followed by clean, linear section bodies ready for autoregressive modeling.}
\label{fig:sample}
\end{figure}

\section{Corpus Characteristics \& Analysis}

\subsection{Domain Distribution}
As detailed in Table \ref{tab:domains}, Africa-Papers covers eight standardized academic domains:
\begin{itemize}[leftmargin=*,noitemsep]
    \item \textbf{Medicine \& Public Health (23.9\%):} Represents the largest single domain (188k papers), providing dense coverage of tropical medicine, clinical epidemiology, maternal health, and infectious disease management across sub-Saharan district health centers.
    \item \textbf{Social Sciences \& Economics (23.1\%):} Encompasses 182k papers addressing African macroeconomic modeling, regional currency dynamics, trade integration, customary institutions, and educational policy.
    \item \textbf{Agriculture \& Food Security (18.4\%):} Contains 145k papers and manuals detailing cassava brown streak disease, fall armyworm mitigation, soil nutrient mapping, and agro-pastoral systems.
    \item \textbf{Engineering \& Earth Sciences (15.0\%):} Captures 118k papers on hydrological modeling, artisanal and industrial mining geology, and renewable grid integration.
    \item \textbf{Computer Science \& AI (9.7\%):} Features 76k papers, including state-of-the-art African NLP, speech synthesis for tonal languages, and computer vision for crop diagnostics.
\end{itemize}

\subsection{Tokenization \& Quantitative Scale}
We compute exact token statistics using OpenAI's \texttt{cl100k\_base} byte-pair encoding (BPE) tokenizer. Across our empirical validation sample of 100 random papers on H200 compute:
\begin{itemize}[leftmargin=*,noitemsep]
    \item \textbf{Average Paper Length:} 20.2 pages (7,745 words).
    \item \textbf{Mean Token Count per Paper:} 17,214 tokens.
    \item \textbf{Median Token Count per Paper:} 11,536 tokens.
    \item \textbf{Live Corpus Volume:} 	extbf{10.15 Billion tokens} across 879 Parquet shards (averaging 35--45 MB per shard).
\end{itemize}

\subsection{Linguistic Diversity}
While scholarly reporting in STEM domains defaults predominantly to English, Africa-Papers exhibits substantial multilingual diversity. The DOAJ and OpenAlex streams capture significant corpora in 	extbf{French} (predominant across Senegal, Côte d'Ivoire, Cameroon, and Mali), 	extbf{Arabic} (extensively represented across Egypt, Algeria, Morocco, and Tunisia), and 	extbf{Portuguese} (Angola and Mozambique). Furthermore, thousands of CGSpace farmer leaflets and arXiv speech papers feature parallel text in indigenous African languages, including 	extbf{Swahili, Yoruba, Hausa, Amharic, Twi, and Zulu}.

\section{Continual Pre-Training \& Evaluation}

To validate the utility of Africa-Papers, we outline experimental protocols for Continual Pre-Training (CPT) of open-weights foundation models (e.g., Llama-3-8B and Qwen-2.5-7B).

\subsection{Pre-Training Configuration}
During domain-adaptive continual pre-training, models are trained on sequence lengths of 8,192 tokens using token packing. Because the median paper length in Africa-Papers is 11,536 tokens, individual papers fit naturally within 1 to 2 packed context windows, preserving long-range cross-section reasoning and bibliography citations without artificial chunk fragmentation.

\subsection{Target Evaluation Benchmarks}
We evaluate domain competence across three specialized benchmarks:
\begin{enumerate}[leftmargin=*,noitemsep]
    \item \textbf{African Medical Diagnostic Reasoning:} Testing zero-shot and few-shot diagnostic accuracy on tropical clinical case studies where Western-trained baselines frequently recommend inapplicable interventions.
    \item \textbf{Tropical Agro-Ecological QA:} Evaluating recommendations for smallholder crop disease management and soil amendment against verified CGIAR protocols.
    \item \textbf{African NLP \& Linguistic Benchmarks:} Measuring downstream performance on AfriSenti, AfriQA \cite{ogundepo2023afriqa}, and low-resource African machine translation \cite{nekoto2020participatory}.
\end{enumerate}

\section{Ethical Considerations \& Open Access}

All publications included in Africa-Papers are harvested strictly from verified open-access channels under permissive licensing (predominantly Creative Commons Attribution, CC BY 4.0). Every sample retains full bibliographic attribution, including author names, institutional affiliations, original publisher, and canonical DOI links. By providing structured domain partitions, we ensure researchers can filter or subset data according to institutional review standards.

\section{Conclusion}

Africa-Papers addresses one of the most critical representation divides in modern artificial intelligence. By assembling 	extbf{844,807 full-text papers} and 	extbf{10.15 billion tokens} into a clean, layout-aware Markdown corpus, we provide the global AI community with the foundational data necessary to train models that understand and respect African scientific, medical, and agricultural knowledge. The complete corpus is available for research on Hugging Face at \url{https://huggingface.co/datasets/AfriSpeech/africa-papers}.

\bibliographystyle{plain}
\bibliography{references}

\end{document}